%% file: main.tex
\documentclass[sigconf, nonacm]{acmart}
\usepackage{colortbl}
\usepackage{multirow}
\usepackage{enumitem}
\usepackage{pifont}
\usepackage{algorithm}
\usepackage{algpseudocode}
\usepackage{pgfplots}
\pgfplotsset{compat=1.18}
\newcommand{\sd}[1]{\textsubscript{\color{gray}\scriptsize #1}}
\definecolor{groupgray}{gray}{0.9}
\definecolor{oursrow}{HTML}{EAF0FB}
\definecolor{heat}{HTML}{3A66C9}
\newcommand{\cmark}{\ding{51}}\newcommand{\xmark}{\ding{55}}
\newcommand{\grouprow}[1]{\rowcolor{groupgray}\multicolumn{13}{c}{\textbf{#1}}\\}
\AtBeginDocument{%
  }

\setcopyright{none}
\renewcommand\footnotetextcopyrightpermission[1]{}

\begin{document}

\title[]{Retrieving from Your Own Mistakes: On-Policy Training of Embedding Models for Multi-Hop Retrieval-Augmented Generation}

\author{Chenjie Qiu}
\affiliation{%
  \institution{Singapore Management University}
  \country{Singapore}}
\email{yausankit@gmail.com}

\author{Zixuan Zeng}
\affiliation{%
  \institution{Guangdong University of Technology}
  \city{Guangzhou}
  \country{China}}
\email{zengzixuan44@mails.gdut.edu.cn}

\author{Shaobin Shi}
\affiliation{%
  \institution{Guangdong University of Technology}
  \city{Guangzhou}
  \country{China}}
\email{d7inshi@gmail.com}

\author{Bingfeng Chen}
\affiliation{%
  \institution{Guangdong University of Technology}
  \city{Guangzhou}
  \country{China}}
\email{chenbingfeng@gdut.edu.cn}

\author{Dongsheng Luo}
\affiliation{%
  \institution{Singapore Management University}
  \country{Singapore}}
\email{dsluo@smu.edu.sg}

\author{Boyan Xu}
\authornote{Corresponding author.}
\affiliation{%
  \institution{Guangdong University of Technology}
  \city{Guangzhou}
  \country{China}}
\email{hpakyim@gmail.com}

\renewcommand{\shortauthors}{Qiu et al.}

\begin{abstract}
\input{sections/abstract}
\end{abstract}


\maketitle

\input{sections/introduction}
\input{sections/related_work}
\input{sections/preliminaries}
\input{sections/method}
\input{sections/experiments}

\input{sections/conclusion}

\bibliographystyle{ACM-Reference-Format}
\bibliography{references}

\appendix
\input{sections/appendix}

\end{document}

%% file: sections/abstract.tex
Complex questions in retrieval-augmented generation require evidence spread
over several documents, and an answer is reliable only when the retriever
finds all of it. Chain embedding retrievers collect such evidence step by
step, re-encoding the question together with the passages retrieved so far,
and we find that they are already strong: trained only on gold histories, a
small chain embedding retriever that calls no language model outperforms
graph-based systems and is competitive with a search agent. This
training, however, leaves a gap. At test time the retriever conditions on its own
retrievals, which soon contain mistakes: by the third step, about half or
more of the states it visits hold a wrong passage, and these passages lower
its chance of retrieving a missing gold passage well beyond what the
difficulty of the states explains. We propose on-policy state training,
which trains the encoder on the states it reaches itself. At every epoch the
current encoder retrieves chains for the training questions; each state it
visits, including those built on its own mistakes, is labeled with the
earliest gold passage still missing while the other missing gold passages
are masked, so that no correct continuation is penalized; and the encoder is
updated contrastively on these states mixed with teacher-forced ones, so
that its training distribution follows it as it learns. The resulting
retriever retrieves complete evidence more often than teacher-forced
training and the systems we compare, and it transfers without further
training to harder long-chain variants of HotpotQA and 2WikiMultiHopQA that
we construct. On the same states, it recovers from its own mistakes far more
often, and controlled comparisons attribute the gain to training on the
states the encoder currently reaches. Code and benchmarks are available at
\url{https://github.com/yausankit/on-policy-state-training}.

%% file: sections/introduction.tex
\section{Introduction}
\label{sec:intro}

\begin{figure*}[t]
  \centering
  \includegraphics[width=\textwidth]{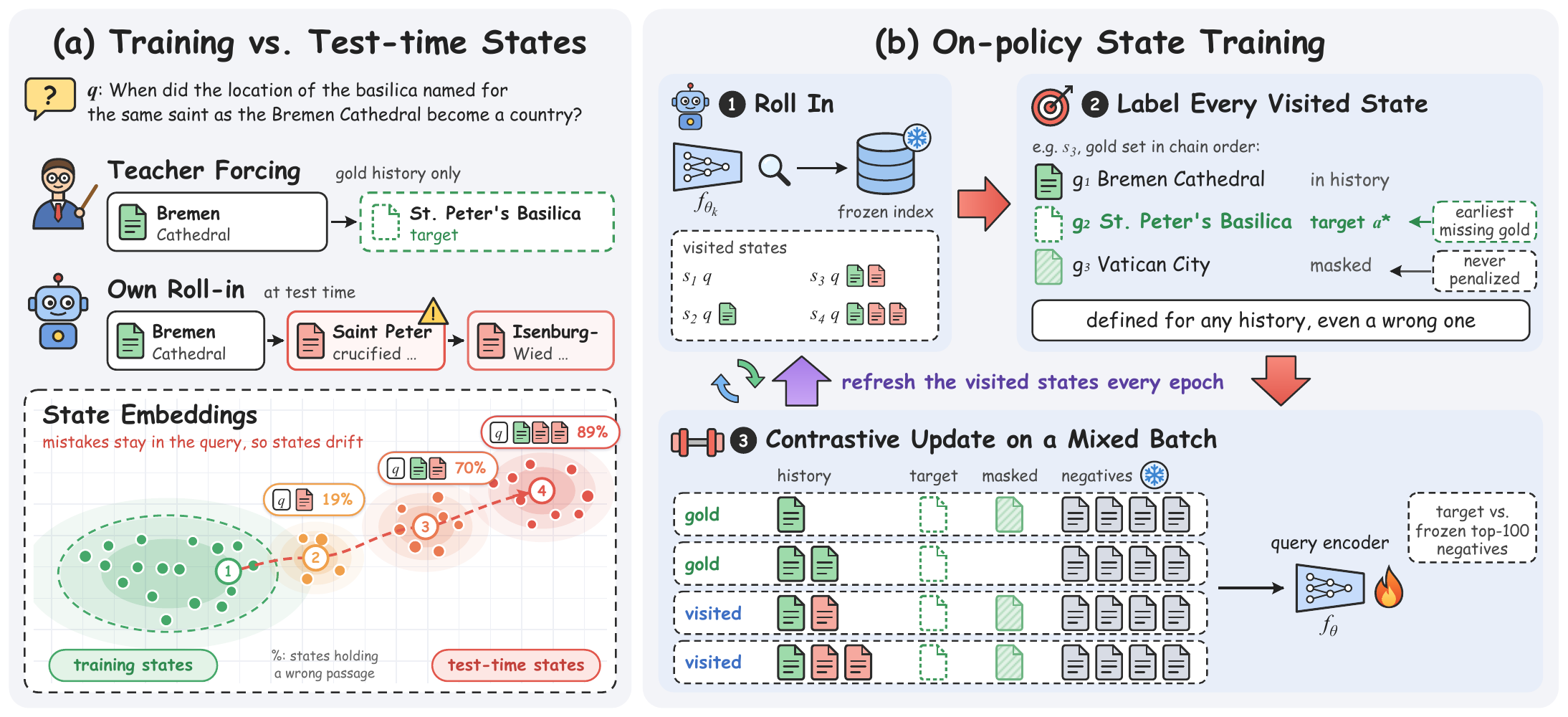}
  \caption{Overview. (a) Teacher forcing trains on gold histories, while at
  test time the retriever's own mistakes stay in the query and its states
  drift. (b) On-policy state training rolls in the current encoder, labels
  every visited state with the earliest missing gold passage, and trains on
  a mix of gold and visited states, refreshed every epoch.}
  \label{fig:overview}
\end{figure*}

Retrieval-augmented generation grounds a language model in retrieved
documents \cite{lewis2020rag}, and on the Web it lets search engines and
assistants answer questions from large collections of pages that change
faster than any model can be retrained. For complex questions the evidence is spread
over several passages, and a missing passage is often enough to make the
answer wrong \cite{trivedi2022musique}. Two families of systems address this
setting. One places a language model in the retrieval loop, either to reason
between searches \cite{jin2025searchr1} or to organize the corpus as a graph
in advance \cite{gutierrez2025hipporag2}. The other trains
an embedding model to retrieve in a chain: at each step it encodes the
question together with the passages found so far and retrieves the next one
\cite{xiong2021mdr}. Chain embedding
retrievers need no language model call and a single dense index, which makes
them a practical first stage for Web-scale corpora, and a better one also helps
systems of the first family.

A chain retriever is trained on one kind of history and used on another.
During training, the passages in its history are gold passages, and the
target is the next gold passage \cite{xiong2021mdr}. At test time the
history consists of whatever the retriever itself retrieved. As soon as one
of these passages is wrong, it stays in the query for every later step, and
the retriever faces a state it was never trained on. Such states are
common. On four-hop MuSiQue questions, 70\% of the states that a
teacher-forced chain retriever visits at its third step, and 89\% at its
fourth, already contain a non-gold passage (Section~\ref{sec:prelim}). In
such states the retriever retrieves a missing gold passage 28\% of the time,
against 67\% when its history is clean. Part of this gap reflects harder
states, but deleting the wrong passages from the same states shows that they
alone cost the retriever 11 points (Section~\ref{sec:rq3}). We call this
the state mismatch. Chain dense
retrievers trained with supervised targets use teacher forcing, and those
that learn from their own trajectories rely on outcome rewards
\cite{sorokin2026qrag}. Work on dense retrieval has improved the
candidate side of training, through harder or cleaner negatives
\cite{xiong2021ance,qu2021rocketqa}, while the query-side histories remain
gold.

This raises two questions. Does the state mismatch limit chain dense
retrieval, so that changing only the states used for training improves it?
And if so, does the improvement come from the states the mismatch points
to, the histories that contain the retriever's own mistakes?

To answer them we propose on-policy state training, which trains the
encoder on the states that it reaches itself. In every epoch the current
encoder retrieves chains for the training questions, and the states it
visits, including those built on its own mistakes, replace half of the
training states. Each visited state is labeled with the earliest gold passage
that is still missing, while the other missing gold passages are masked so
that retrieving any of them is never penalized; this target is defined by
the gold set alone, so states with wrong passages receive the same dense
supervision as gold states. Because the visited states are re-collected as
the encoder changes, the training distribution follows the retriever as it
learns. Everything else, from the architecture and the loss to the
negatives and the document index, is shared with teacher-forced training,
which isolates the distribution of training states. It also differs
from methods that learn retrieval from outcome rewards
\cite{sorokin2026qrag,jin2025searchr1}, which need sparse rewards over
whole episodes. We evaluate on MuSiQue and, to test
whether the effect persists on longer chains and larger corpora, on harder
variants of HotpotQA and 2WikiMultiHopQA that we construct
\cite{yang2018hotpotqa,ho2020twowiki}.

Our study leads to four findings.

(1) \textbf{A trained chain embedding retriever is already a strong retriever for complex questions.} Even with teacher forcing, a 0.6B chain retriever that calls no language model retrieves complete evidence more often than graph-based retrieval-augmented generation and is competitive with a 7B search agent, at a fraction of their retrieval cost.

(2) \textbf{The state mismatch still limits it.} Changing only the training states raises complete coverage over teacher forcing with the same encoder on MuSiQue and on both long-chain benchmarks, the resulting retriever outperforms every system we compare on MuSiQue, and the MuSiQue-trained encoder transfers to longer chains.

(3) \textbf{The gain appears where the mismatch predicts.} Scored on the same states, teacher-forced and on-policy encoders continue a chain from a clean history about equally well, while after one or more mistakes the on-policy encoder recovers far more often, so the gain accumulates along the chain.

(4) \textbf{Exposure to contaminated histories explains part of the gain, and following the current encoder explains the rest.} Synthetic contamination of gold histories, matched in the number of wrong passages, closes about half of the gap to on-policy state training. Visited states collected once are worth little more than synthetic contamination, and the remaining gain comes from refreshing them as the encoder changes; resampling synthetic contamination every epoch does not reproduce it (Section~\ref{sec:rq4}).

%% file: sections/related_work.tex
\section{Related Work}
\label{sec:related}

\subsection{Retrieval for Complex Questions}
Questions that require several pieces of evidence are usually answered by
retrieving in multiple rounds. One line of work places a large language model
in the loop and learns when and what to search through reinforcement
learning \cite{jin2025searchr1}. Graph-based methods organize the corpus
offline instead, as a knowledge graph traversed with personalized PageRank
\cite{gutierrez2025hipporag2}, a subgraph selected with a prize-collecting
Steiner tree \cite{he2024gretriever}, or a relation-free entity graph
\cite{zhuang2026linearrag}. Our method calls no language model at retrieval
time and uses only a dense passage index. The two directions are
complementary, since a stronger embedding model can serve as the retriever
inside either kind of system. An embedding retriever can also be trained
from the outcome of a whole episode: Q-RAG learns a value-based embedder for
multi-step retrieval from temporal-difference targets \cite{sorokin2026qrag}.
Search-R1 and Q-RAG learn from sparse outcome rewards. Our method derives a dense
per-step target from the gold evidence set, so every visited state is
labeled without reward estimation, and neither training nor retrieval calls
a language model.

\subsection{Chain and Dense Retrievers}
Chain dense retrievers condition each hop on the passages retrieved at
earlier hops. MDR concatenates the question with the previously retrieved
passages and encodes the result with a shared dense encoder, trained with
teacher forcing: the history at hop $t$ consists of gold passages, the target
is the next gold passage, and negatives come from a memory bank of passage
vectors \cite{xiong2021mdr}. Its training queries therefore never contain a
passage the retriever retrieved by mistake, and no target is defined for a
history that does. We keep the MDR-style chain architecture and a fixed
document index, and change only the distribution of histories on which the
query encoder is trained.

Dense retrievers are trained contrastively on question--passage pairs
\cite{karpukhin2020dpr}, and much of the progress comes from better negatives.
ANCE re-encodes the corpus with the current checkpoint so that negatives
follow the model's own retrieval distribution \cite{xiong2021ance}, and
RocketQA removes false negatives with a cross-encoder \cite{qu2021rocketqa}.
Contriever and recent embedding models built on large language models scale
up pretraining
\cite{izacard2022contriever,lee2024nvembed,zhang2025qwen3embedding}. These
methods improve the candidate side of the contrastive objective. Our method
acts on the query side: it refreshes the retrieval states that the encoder
embeds as queries, while the negatives stay those of the frozen encoder.

\subsection{Exposure Bias and On-Policy Training}
Training a sequential predictor only on reference histories, while it
conditions on its own predictions at test time, is known as exposure bias
\cite{ranzato2016sequence}. Imitation learning and sequence modeling address it
by training on states the model itself visits
\cite{daume2009searn,ross2011dagger,bengio2015scheduled}. DAgger rolls in
with the learner and labels the visited states with an expert, and dynamic
oracles define such an expert for every state of a dependency parser,
including states off the gold path \cite{goldberg2012dynamic}; AggreVaTe and
LOLS label states with costs to go instead
\cite{ross2014aggrevate,chang2015lols}. The idea has recently been applied
to language model agents \cite{li2026daggerllm,liao2026reopd}. Our training procedure is an instance of DAgger with a
dynamic oracle. What we study is specific to chain dense retrieval: the
oracle is free because the gold evidence set labels every state, a wrong
passage persists in the query for all later steps, and only the query side
of a contrastive objective changes. A related line makes language models
robust to irrelevant retrieved context by training with distractors
\cite{yoran2024robust,zhang2024raft}; Section~\ref{sec:rq4} compares
visited states with synthetic contamination of this kind.

%% file: sections/preliminaries.tex
\section{Problem Setup}
\label{sec:prelim}

\subsection{Chain Dense Retrieval}
A question $q$ is paired with a corpus $\mathcal{C}$ and a set of gold
passages $G(q)=\{g_1,\dots,g_n\}\subseteq\mathcal{C}$, listed in the order in
which a reasoning chain reaches them. A chain retriever collects evidence over
$T$ steps. Its state at step $t$ is $s_t=(q,E_t)$, where
$E_t=(e_1,\dots,e_{t-1})$ holds the passages it has retrieved so far and
$E_1=\emptyset$. Following MDR \cite{xiong2021mdr}, the state is serialized as
a single text $x(s)$ that lists the question and the retrieved passages, and
is embedded by a query encoder $f_\theta$. Passages are embedded once by a
fixed document encoder $h$. With both embeddings normalized, the retriever
scores a passage $d$ in state $s$ by
\begin{equation}
  \sigma_\theta(s,d)=f_\theta\big(x(s)\big)^{\top}h(d),
\end{equation}
and its greedy policy $\pi_\theta$ retrieves the highest-scoring passage it has
not retrieved yet,
\begin{equation}
  e_t=\operatorname*{arg\,max}_{d\in\mathcal{C}\setminus E_t}\sigma_\theta(s_t,d),
  \qquad s_{t+1}=(q,E_t\oplus e_t).
\end{equation}
The output ranking lists $e_1,\dots,e_T$ first, followed by the remaining
passages ordered by their best score over the $T$ steps,
$\max_t\sigma_\theta(s_t,d)$. We measure recall R@$k$, the fraction of $G(q)$
in the top $k$, and complete coverage C@$k$, the fraction of questions whose
whole gold set is in the top $k$. C@$k$ reflects whether a downstream
retrieval-augmented generation system receives all of the evidence it needs.

\subsection{Supervision from the Gold Set}
For any state $s=(q,E)$, the gold passages still missing are
$R(s)=G(q)\setminus E$, and the target is the earliest of them in chain order,
\begin{equation}
  a^*(s)=g_j,\qquad j=\min\{\,i : g_i\in R(s)\,\}.
\end{equation}
The target depends only on $G(q)$ and on which passages $E$ contains, so it is
defined for every state with $R(s)\neq\varnothing$, including states whose history contains passages
outside $G(q)$. Given a set of candidate negatives $N(s)$, the retriever is
trained with a contrastive loss with temperature $\tau$,
\begin{equation}
  \ell_\theta(s)=-\log
  \frac{\exp\big(\sigma_\theta(s,a^*(s))/\tau\big)}
       {\sum_{d\in\{a^*(s)\}\cup\big(N(s)\setminus R(s)\big)}
        \exp\big(\sigma_\theta(s,d)/\tau\big)}.
  \label{eq:loss}
\end{equation}
Gold passages that are still missing but are not the current target are
removed from the denominator, since retrieving them is also progress.

\subsection{Teacher Forcing and the State Mismatch}
Existing chain retrievers minimize Eq.~\eqref{eq:loss} over a distribution
$d_{\mathrm{gold}}$ of states whose histories contain only gold passages
\cite{xiong2021mdr}:
\begin{equation}
  \mathcal{J}_{\mathrm{gold}}(\theta)=
  \mathbb{E}_{s\sim d_{\mathrm{gold}}}\big[\ell_\theta(s)\big].
\end{equation}
In MDR-style training $E$ is a gold prefix $(g_1,\dots,g_k)$. A variant that
we also study draws $E$ as a random subset of $G(q)$ in chain order, which
covers histories that skip a hop. At test time the retriever instead
conditions on the states $d_{\pi_\theta}$ that its own policy visits. Once
$\pi_\theta$ writes a passage outside $G(q)$, every later state lies outside
the support of $d_{\mathrm{gold}}$, and the wrong passage stays in the query
text $x(s)$ for all remaining steps. Teacher-forced training therefore never
shows the encoder the histories in which it most needs to recover. This is
the exposure bias studied in sequence prediction and imitation learning
\cite{ranzato2016sequence,daume2009searn,ross2011dagger}. The rest of the
paper asks how much this mismatch limits chain dense retrieval, and which
properties of the visited states matter when it is corrected.

\subsection{How Large Is the Mismatch?}
We measure it with a chain retriever trained by teacher forcing on MuSiQue
(gold histories drawn as random subsets), run for $T=4$ steps on the 1,000
test questions. We count only feasible states: those at step $t\le n$
of an $n$-hop question that still miss a gold passage, since for $t>n$ a
missing gold passage implies a wrong one. On four-hop questions, 19\% of the
states at the second step, 70\% at the third and 89\% at the fourth already
contain a non-gold passage; on three-hop questions the share is 12\% at the
second step and 48\% at the third (Figure~\ref{fig:diagnosis}a). The
retriever retrieves a missing gold passage in 67\% of the clean states and in
28\% of the contaminated ones. The gap is not explained by question length:
it is 71\% against 33\% on two-hop questions, 67\% against 40\% on three-hop
questions and 57\% against 19\% on four-hop questions. Part of this gap reflects
harder states. A counterfactual separates the two: deleting the wrong passages from the same contaminated
states raises the success rate of the teacher-forced encoder from 20.8\% to
32.1\%, so its own mistakes cost it about a third of its remaining chance of
continuing the chain. Teacher forcing trains the encoder only on clean
states, while a large share of the states it meets after the second step are
of the other kind.

%% file: sections/method.tex
\section{On-Policy State Training}
\label{sec:method}

To study the effect of the state mismatch, we change a single factor of
teacher-forced training: the distribution of the states on which the query
encoder is trained. The architecture, the target $a^*(s)$, the loss in
Eq.~\eqref{eq:loss}, the negative candidates and the document index all stay
as they are, so any change in retrieval can be attributed to which states
the encoder sees during training. The procedure follows DAgger
\cite{ross2011dagger} with a dynamic oracle \cite{goldberg2012dynamic}: the
learner rolls in, and an oracle defined on every state, including states off
the gold path, provides the label (Figure~\ref{fig:overview}b). In chain dense retrieval this oracle
comes from the gold evidence set at no cost, and the design choices below
adapt the procedure to a contrastive query encoder over a fixed index.

\subsection{Objective}
We train on a mixture of teacher-forced states and states visited by the
current retriever,
\begin{equation}
  \mathcal{J}(\theta)=
  \mathbb{E}_{s\sim\mu_\theta}\big[\ell_\theta(s)\big],\qquad
  \mu_\theta=\lambda\, d_{\mathrm{gold}}+(1-\lambda)\, d_{\pi_\theta},
  \label{eq:objective}
\end{equation}
with $\lambda=\tfrac12$. Since $d_{\pi_\theta}$ depends on $\theta$, we treat
visited states as data and alternate between collecting them and fitting
them. At the start of epoch $k$, the current encoder $\theta_k$ runs its
greedy policy for $T$ steps on every training question. Each visited state
with a non-empty $R(s)$ is recorded, duplicates of the same $(q,E)$ are
removed, and $M$ of the remaining states are sampled uniformly to form
$\mathcal{D}^{\pi}_k$. The epoch then minimizes Eq.~\eqref{eq:loss} over
\begin{equation}
  \mathcal{D}_k=\mathcal{D}^{\mathrm{gold}}\cup\mathcal{D}^{\pi}_k ,
\end{equation}
where $\mathcal{D}^{\mathrm{gold}}$ consists of $M$ teacher-forced
states drawn as random gold subsets, and $M$ is half of the about 4,700 training
states of teacher forcing. The LoRA parameters are initialized
from scratch, as for teacher forcing; in the first epoch the visited states
come from a roll-in of a teacher-forced encoder, which only supplies states
and is not used to initialize the model. Both regimes use the same number of
states and the same schedule of at most 10 epochs, with early stopping on the
validation questions. The visited half is replaced at every
epoch instead of being accumulated across epochs as in DAgger
\cite{ross2011dagger}, so it follows the states that the encoder currently
reaches. The teacher-forced half keeps supervision on histories that the
current encoder seldom reaches, such as long correct prefixes on questions
it usually gets wrong, which would otherwise disappear from training. We
refer to this procedure as on-policy state training.

\subsection{Labels for Visited States}
Visited states can be labeled at no extra cost because $a^*(s)$ is defined by
the gold set alone. When the history contains a wrong passage, the target is
still the earliest gold passage not yet retrieved, and the other missing gold
passages are removed from the denominator of Eq.~\eqref{eq:loss}, so the encoder is trained
to continue the chain from a contaminated history. Every retained visited state
receives a dense, per-step target. No language model, reward model or
outcome reward is involved, which distinguishes this setting from retrievers
trained with reinforcement learning
\cite{sorokin2026qrag,jin2025searchr1}. Because the target
function is shared with teacher forcing, the two training regimes differ only
in their state distributions. The rule fixes which missing gold passage serves as the
positive, but since the other missing gold passages are masked, retrieving
any of them is never penalized; on questions whose chain order is partly
arbitrary, such as comparisons, the ordering only selects which correct
continuation receives the gradient.

\subsection{Fixed Index and Negative Rule}
Only the query encoder is updated, through low-rank adapters
\cite{hu2022lora}, and the document embeddings $h(d)$ are computed once. The
negatives $N(s)$ for every state are drawn from the top 100 passages that the
frozen encoder retrieves for the state text $x(s)$, excluding every gold
passage of the question, whether the state is teacher-forced or visited.
Non-gold passages already in a visited history can therefore appear among its
negatives.
We fix both for the comparison. If the corpus were re-encoded, or the
negatives were mined with the current model as in ANCE
\cite{xiong2021ance}, a gain could come from harder negatives instead of
from the states. With both held fixed, the negatives of every state are mined
by the same rule and the same model in both training regimes. The pools
still depend on the state text, so a visited state receives other negatives
than a gold state of the same question; the two regimes share the rule
that chooses them.

%% file: sections/experiments.tex
\section{Experiments}

\label{sec:exp}

We answer four research questions.
\textbf{RQ1}: Does on-policy state training retrieve complete evidence
chains more often than existing retrievers, and does it transfer to longer
chains? \textbf{RQ2}: Which components of the method contribute?
\textbf{RQ3}: Does the gain come from recovering after the retriever's own
mistakes? \textbf{RQ4}: What in the visited states matters?

\subsection{Experimental Setup}

\subsubsection{Datasets}
MuSiQue \cite{trivedi2022musique} is built so that its two- to four-hop
questions cannot be answered from a single passage, which makes the history
of a chain retriever matter at every step. All trained retrievers are trained on 1,800 MuSiQue training questions and
select checkpoints on 200 others. We test on the 1,000 questions of the
official development set with the 11,656-passage corpus used by HippoRAG~2
\cite{gutierrez2025hipporag2}. To
test whether the effect persists on longer chains and larger corpora, we
evaluate the MuSiQue-trained encoders, without further training, on two
harder benchmarks that we construct from HotpotQA \cite{yang2018hotpotqa}
and 2WikiMultiHopQA \cite{ho2020twowiki}. HotpotQA-Long extends bridge
questions along hyperlinks in the Wikipedia abstracts to chains of three,
four and five pages (100 test questions each) over a corpus of 554,812
passages. 2Wiki-Long follows functional relations between Wikidata entities
for three to five hops, or compares three to four entities that are each
reached in two hops, with 300 test questions over 220,000 passages. In both
benchmarks the question names only the starting entity, a language model of
a different family verifies that the question is answerable from the gold
passages and has a unique answer, and questions whose final passage is
retrievable by BM25 from the question alone are removed. The corpora contain
the lexical nearest neighbors of every gold passage, so distractors are
topically close to the evidence (Appendix~\ref{app:data}).

\subsubsection{Metrics}
With $L_k(q)$ the top $k$ passages of the ranking for question $q$ and
$\mathcal{Q}$ the test questions, we report recall and complete coverage for
$k\in\{5,10\}$:
\begin{align}
  \mathrm{R@}k&=\frac{1}{|\mathcal{Q}|}\sum_{q\in\mathcal{Q}}
  \frac{|G(q)\cap L_k(q)|}{|G(q)|},\\
  \mathrm{C@}k&=\frac{1}{|\mathcal{Q}|}\sum_{q\in\mathcal{Q}}
  \mathbf{1}\big[G(q)\subseteq L_k(q)\big].
\end{align}
C@$k$ credits a question only when every gold passage is retrieved, which is
what a downstream reader needs.

\subsubsection{Baselines}
BM25 and the frozen Qwen3-Embedding-0.6B encoder
\cite{zhang2025qwen3embedding} retrieve with the question alone (Dense).
HippoRAG~2 \cite{gutierrez2025hipporag2} builds a knowledge graph with a
language model and retrieves with personalized PageRank; we run the
official code with the same frozen encoder. G-Retriever
\cite{he2024gretriever} selects a subgraph of this knowledge graph with a
prize-collecting Steiner tree; we rank the passages linked to the selected
nodes and edges first. LinearRAG \cite{zhuang2026linearrag} builds an
entity graph without relation extraction and retrieves by activation
spreading over it. Both use the same frozen encoder. The released MDR checkpoint
\cite{xiong2021mdr} is a chain retriever trained with teacher forcing on
HotpotQA; it retrieves for two hops with beam search.
Search-R1 \cite{jin2025searchr1} is a 7B language model trained with
reinforcement learning to issue search queries. Q-RAG
\cite{sorokin2026qrag} trains a multi-step embedding retriever with
value-based reinforcement learning; we evaluate the authors' released
MuSiQue checkpoint with its own encoder and default four steps. For
Search-R1 we use the released
checkpoint with its default settings (three passages per search, at most
four searches) and our frozen encoder as its search engine, and rank
passages in the order they are retrieved. Our retriever uses the same
encoder, adapted with LoRA on the query side only, and runs $T=4$ chain
steps on MuSiQue and $T=8$ on the long-chain benchmarks. Table~\ref{tab:main} averages five
seeds; all other experiments train each retriever once with one shared
seed. All models are trained and evaluated on NVIDIA RTX 4090 GPUs;
Appendix~\ref{app:details} gives the training details and
Appendix~\ref{app:cost} the training and retrieval cost.

\begin{table*}[t]
  \centering
  \caption{Retrieval results. Our retriever is trained on MuSiQue and
  evaluated on the long-chain benchmarks without further training; its
  results are the mean over five seeds, with the standard deviation in gray. Dense is the frozen Qwen3-Embedding-0.6B encoder with
  the question as the query; the structure-augmented baselines, Search-R1 and
  our retriever use the same encoder. Search-R1 calls a 7B language model at every step. The best
  result in each column is in bold and the second best is underlined. $^\dagger$HippoRAG~2
  and G-Retriever on the long-chain benchmarks are run on reduced corpora
  (all gold passages plus the frozen encoder's top 100 per question), an
  easier setting than the full corpora used by the other rows
  (Appendix~\ref{app:pool}).}
  \label{tab:main}
  \setlength{\tabcolsep}{3.6pt}
  \begin{tabular}{lcccccccccccc}
    \toprule
    \multirow{2}{*}{Method} & \multicolumn{4}{c}{MuSiQue} & \multicolumn{4}{c}{HotpotQA-Long}
      & \multicolumn{4}{c}{2Wiki-Long} \\
    \cmidrule(lr){2-5}\cmidrule(lr){6-9}\cmidrule(lr){10-13}
    & R@5 & R@10 & C@5 & C@10 & R@5 & R@10 & C@5 & C@10 & R@5 & R@10 & C@5 & C@10 \\
    \midrule
    \grouprow{Simple Baselines}
    BM25 & 43.4 & 50.8 & 9.8 & 15.5 & 23.3 & 25.5 & 0.0 & 0.0 & 27.9 & 31.5 & 0.0 & 0.0 \\
    Dense & 55.1 & 62.4 & 18.9 & 27.1 & 29.0 & 31.9 & 0.3 & 1.0 & 31.3 & 34.5 & 0.0 & 2.0 \\
    \grouprow{Structure-Augmented Retrieval}
    G-Retriever & 36.5 & 43.7 & 9.2 & 15.9 & 16.1$^\dagger$ & 20.5$^\dagger$ & 0.0$^\dagger$ & 2.0$^\dagger$ & 18.9$^\dagger$ & 22.2$^\dagger$ & 0.0$^\dagger$ & 1.7$^\dagger$ \\
    HippoRAG 2 & 58.2 & 66.0 & 23.1 & 32.3 & 30.9$^\dagger$ & 33.4$^\dagger$ & 1.0$^\dagger$ & 1.3$^\dagger$ & 32.1$^\dagger$ & 35.7$^\dagger$ & 0.0$^\dagger$ & 2.3$^\dagger$ \\
    LinearRAG & 43.9 & 55.8 & 17.7 & 28.7 & 19.0 & 22.1 & 0.7 & 0.7 & 23.2 & 29.3 & 0.3 & 1.0 \\
    \grouprow{Sequential Retrieval}
    MDR & 58.3 & 66.0 & 31.9 & 38.3 & \textbf{51.0} & 56.2 & \underline{11.0} & 14.7 & 46.9 & 51.5 & \underline{10.0} & 14.3 \\
    Q-RAG & 61.7 & 67.8 & 36.2 & 44.9 & 25.2 & 28.0 & 3.3 & 5.7 & 14.7 & 27.8 & 1.7 & 2.3 \\
    Search-R1 & \underline{66.7} & \underline{74.7} & \underline{36.5} & \underline{49.5}
      & 43.6 & \textbf{58.5} & 3.3 & \underline{25.7}
      & \underline{50.3} & \textbf{68.1} & 6.0 & \textbf{37.0} \\
    \midrule
    \textbf{On-policy state training} & \textbf{73.5}\sd{0.4} & \textbf{80.5}\sd{0.5} & \textbf{49.3}\sd{0.4} & \textbf{59.8}\sd{0.9}
      & \underline{49.6}\sd{1.3} & \underline{56.6}\sd{0.7} & \textbf{20.8}\sd{2.1} & \textbf{29.6}\sd{0.9}
      & \textbf{59.2}\sd{2.9} & \underline{66.1}\sd{2.9} & \textbf{27.7}\sd{4.5} & \underline{32.8}\sd{5.2} \\
    \bottomrule
  \end{tabular}
\end{table*}

\subsection{RQ1: Retrieval Effectiveness}
\label{sec:main}

\subsubsection{Overall Results}

Table~\ref{tab:main} supports the first finding. On MuSiQue, on-policy state
training reaches 73.5 R@5 and 49.3 C@5. Search-R1, which reasons with a 7B
language model between three searches on average, completes 36.5\% of the
questions within five passages, Q-RAG 36.2\%, the released MDR checkpoint
31.9\%, and
HippoRAG~2, which calls a language model to build its graph, 23.1\%.
Most of the gain over the frozen encoder comes from training a chain
retriever at all; the part due to the training states is isolated in
Section~\ref{sec:ablation}, where teacher forcing with the same encoder and
recipe reaches 44.6 C@5 and on-policy state training 50.5. Teacher forcing
alone thus already completes more chains on MuSiQue than every compared
system (44.6 against 36.5 C@5 for Search-R1), and on the long-chain
benchmarks it is far ahead of the graph-based systems and close to
Search-R1 on HotpotQA-Long: a trained chain embedding retriever, which calls no language
model at retrieval time, is a strong retriever for complex questions. The gain in complete coverage is larger than the gain
in recall, which reflects what a chain needs: a question counts toward C@5
only when every step succeeds.

\subsubsection{Transfer to Longer Chains}
MuSiQue has at most four hops, while the long-chain benchmarks need up to
five hops, or six to eight passages for comparisons, over corpora 19 to 48
times larger. The MuSiQue-trained encoder raises C@10 over the frozen
encoder from 1.0 to 29.6 on HotpotQA-Long and from 2.0 to 32.8 on
2Wiki-Long. Search-R1 is the strongest competitor here. It puts fewer gold
passages in its first five (C@5 3.3 and 6.0, against 20.8 and 27.7) but
reaches a similar recall by the tenth passage, and leads on 2Wiki-Long in
C@10. Its policy was trained with reinforcement learning on HotpotQA
questions, from which part of HotpotQA-Long is built, and it spends a 7B
forward pass per search. Our retriever uses a 0.6B encoder and one nearest
neighbor search per step. The released MDR checkpoint, trained on HotpotQA,
retrieves the most gold passages in the top five on HotpotQA-Long (51.0
R@5), but with two hops it completes few of the three- to five-hop chains
(14.7 C@10, against 29.6).

\subsubsection{Results by Question Type}

\begin{figure*}[t]
  \centering
  \includegraphics[width=\textwidth]{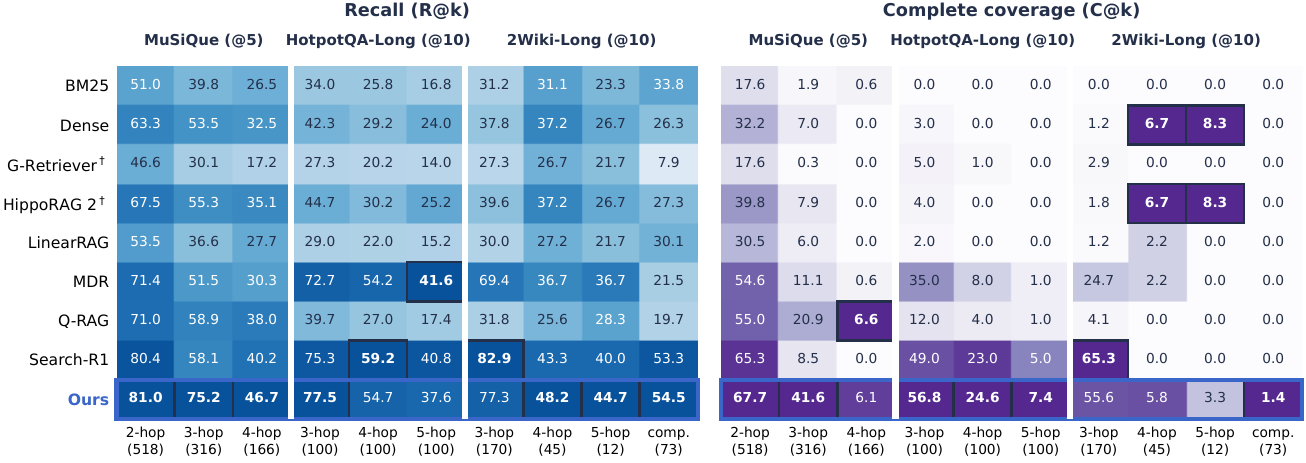}
  \caption{Retrieval by question type, with the number of questions under each
  column. Colors are scaled within each column, and the best result of each
  column is framed. Our retriever is the mean over the five seeds of
  Table~\ref{tab:main}; $^\dagger$reduced corpora.}
  \label{fig:bytype}
\end{figure*}

Figure~\ref{fig:bytype} splits the results of Table~\ref{tab:main} by
question type. On MuSiQue, the gain is largest on three-hop questions, where
our retriever completes 41.6\% of the chains within five passages and the
next system 20.9\%; on four-hop questions every retriever completes fewer
than 7\%. On the long-chain benchmarks, our retriever has the highest
complete coverage on every type of HotpotQA-Long and on the comparison
questions of 2Wiki-Long. On 2Wiki-Long, Search-R1 leads on the three-hop
chains, the closest to the two-hop questions its policy was trained on, and
on the four- and five-hop chains, with 45 and 12 questions, no retriever
completes more than one chain in ten.

\subsubsection{Downstream Question Answering}
Complete coverage matters because a reader rarely answers correctly without the missing
evidence. Table~\ref{tab:qa} gives the top five passages of each retriever
to the same reader (deepseek-flash, temperature 0) and scores its answers on
MuSiQue; Search-R1 is scored on the passages it retrieves, not on its own
answers. The ranking of the retrievers carries over to the answers. With the
passages of on-policy state training, the reader reaches 44.1 exact match,
8.6 points more than with the passages Search-R1 retrieves and 11.7 more
than with those of Q-RAG.

\begin{table}[t]
  \centering
  \caption{Question answering on MuSiQue with a fixed reader; trained retrievers use the shared seed.}
  \label{tab:qa}
  \begin{tabular}{lcc}
    \toprule
    Passages from & EM & F1 \\
    \midrule
    Dense & 25.9 & 36.3 \\
    G-Retriever & 21.2 & 29.2 \\
    HippoRAG 2 & 28.2 & 39.4 \\
    LinearRAG & 22.3 & 30.5 \\
    Q-RAG & 32.4 & 42.0 \\
    Search-R1 & 35.5 & 44.8 \\
    On-policy state training & \textbf{44.1} & \textbf{53.7} \\
    \bottomrule
  \end{tabular}
\end{table}

\subsection{RQ2: Ablations}
\label{sec:ablation}

Table~\ref{tab:ablation} removes the components of the method in turn. Each row
removes one more component, so the last row also has no visited states.
Removing the refresh keeps the visited states collected at the
start of training. Removing the visited states leaves teacher forcing on
gold states alone. Removing the chain fine-tunes the same encoder on the
question alone, which also removes the notion of a retrieval state.

\begin{table}[t]
  \centering
  \caption{Ablations; all rows are trained with the same seed.}
  \label{tab:ablation}
  \setlength{\tabcolsep}{2.4pt}
  \begin{tabular}{lcccccc}
    \toprule
    & \multicolumn{2}{c}{MuSiQue} & \multicolumn{2}{c}{HotpotQA-L} & \multicolumn{2}{c}{2Wiki-L} \\
    \cmidrule(lr){2-3}\cmidrule(lr){4-5}\cmidrule(lr){6-7}
    Method & R@5 & C@5 & R@10 & C@10 & R@10 & C@10 \\
    \midrule
    On-policy state training & \cellcolor{heat!38}\textbf{74.2} & \cellcolor{heat!38}\textbf{50.5} & \cellcolor{heat!38}\textbf{56.4} & \cellcolor{heat!38}\textbf{28.0} & \cellcolor{heat!38}\textbf{67.0} & \cellcolor{heat!38}\textbf{34.7} \\
    \quad w/o refresh & \cellcolor{heat!24}71.8 & \cellcolor{heat!24}48.4 & \cellcolor{heat!24}52.6 & \cellcolor{heat!24}22.7 & \cellcolor{heat!24}64.6 & \cellcolor{heat!24}28.3 \\
    \quad w/o on-policy states & \cellcolor{heat!14}69.9 & \cellcolor{heat!14}44.6 & \cellcolor{heat!14}51.5 & \cellcolor{heat!14}22.3 & \cellcolor{heat!14}61.3 & \cellcolor{heat!14}24.3 \\
    \quad w/o chain & \cellcolor{heat!7}54.6 & \cellcolor{heat!7}18.9 & \cellcolor{heat!7}29.8 & \cellcolor{heat!7}1.0 & \cellcolor{heat!7}31.4 & \cellcolor{heat!7}1.7 \\
    \bottomrule
  \end{tabular}
\end{table}

Every component contributes, and the contributions grow on the longer
chains. Removing the visited states, which leaves teacher forcing with the
same encoder, recipe and number of states, costs 5.9 points of C@5 on
MuSiQue and 5.7 and 10.4 points of C@10 on HotpotQA-Long and 2Wiki-Long. This
comparison isolates the training states, and the gap does not shrink on
chains longer than any seen in training.
Visited states collected once and never refreshed keep most of the gain on
MuSiQue but little on HotpotQA-Long and less than half on 2Wiki-Long
(22.7 against 22.3 and 28.3 against 24.3 C@10), so following the states of the current encoder matters
most when the encoder is run far from its training distribution. Without a
chain, fine-tuning on the question alone gives no gain over the frozen
encoder (54.6 against 55.1 R@5 on MuSiQue), which confirms that complex
retrieval needs the history.

\subsection{RQ3: Recovery from Own Mistakes}
\label{sec:rq3}

The diagnosis in Section~\ref{sec:prelim} predicts where on-policy state
training helps most: the states whose history already holds a wrong
passage. We test that prediction on MuSiQue with the teacher-forced,
collected-once and refreshed retrievers of Table~\ref{tab:decomp}. Comparing each retriever on the
states it visits itself would confound the effect of training with the
states each one happens to reach, so the probes below score all retrievers
on one fixed set of states.

\begin{figure}[t]
  \centering
  \includegraphics[width=\columnwidth]{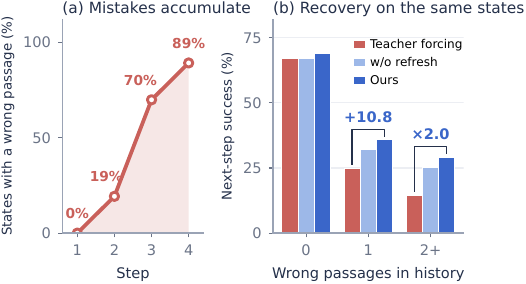}
  \caption{(a) Share of the teacher-forced retriever's states whose history
  contains a wrong passage, at each step of four-hop MuSiQue questions (gold
  passage still missing). (b) Rate at which each encoder retrieves a missing
  gold passage next on the same states, by the number of wrong passages in
  the history. All encoders are trained with the same seed.}
  \label{fig:diagnosis}
\end{figure}

\subsubsection{Recovery on Fixed States}
We take the 3,403 states visited by the teacher-forced retriever that still
miss a gold passage and let every encoder choose the next passage for them
(Figure~\ref{fig:diagnosis}b). On its own states, the teacher-forced encoder
reproduces 99.8\% of its recorded choices, so the probe matches retrieval.
If on-policy state training only made the encoder better at retrieval in
general, the gain would appear in clean states as well. It barely does: with
a clean history, the three encoders succeed in 67.0\%, 67.2\% and 69.0\% of
the states. After one wrong passage, the rate rises from 25.0\% under
teacher forcing to 35.8\% under on-policy state training, and after two or
more from 14.5\% to 29.2\%, twice as often. Visited states collected once,
without refresh, recover most of the way (32.2\% and 25.1\%). Because all
encoders see the same states, the difference lies in the encoders
themselves. The same pattern holds on the 3,352 states visited by the
on-policy retriever: 66.8\% against 68.5\% with a clean history, 26.1\%
against 33.5\% after one wrong passage and 20.0\% against 24.2\% after two
or more.

\subsubsection{Deleting the Wrong Passages}
To measure the effect of the contamination itself, we delete the non-gold
passages from the contaminated states above and keep the gold ones in order;
the deleted passages stay excluded from retrieval, as in the original state.
On these 1,358 states, the teacher-forced encoder succeeds in 20.8\% of the
original states and in 32.1\% once the wrong passages are removed: its own
mistakes cost it 11.3 points. The on-policy encoder succeeds in 33.1\% and
35.3\%, so the same contamination costs it 2.2 points. Part of the low success after a mistake reflects harder states, since
even the cleaned states stay well below the 67\% of states that were clean to
begin with. The rest is the effect of the wrong passages in the query, and
on-policy state training removes most of it: on the contaminated states, the
on-policy encoder (33.1\%) does slightly better than the teacher-forced
encoder does once an oracle has deleted the wrong passages (32.1\%). It is
also better on the cleaned states themselves (35.3\% against 32.1\%), which
are gold histories with gaps, so part of the gain is skill on histories that
skip a hop.

\subsubsection{Gain along the Chain}
Because the encoder recovers more often, the gain accumulates over steps.
The rate of retrieving a gold passage changes by one to two points at the
first two steps, where most histories are still clean, and rises from 27.7\%
to 38.7\% at the third step and from 15.2\% to 23.1\% at the fourth. By
question length, C@5 rises from 61.6 to 67.8 on two-hop questions and from
36.7 to 44.3 on three-hop questions. On four-hop questions recall rises
from 44.1 to 48.3 R@5, but complete coverage within five passages stays
below 10\% for both retrievers, since it requires four consecutive
successful steps.

\begin{figure*}[t]
  \centering
  \includegraphics[width=\textwidth]{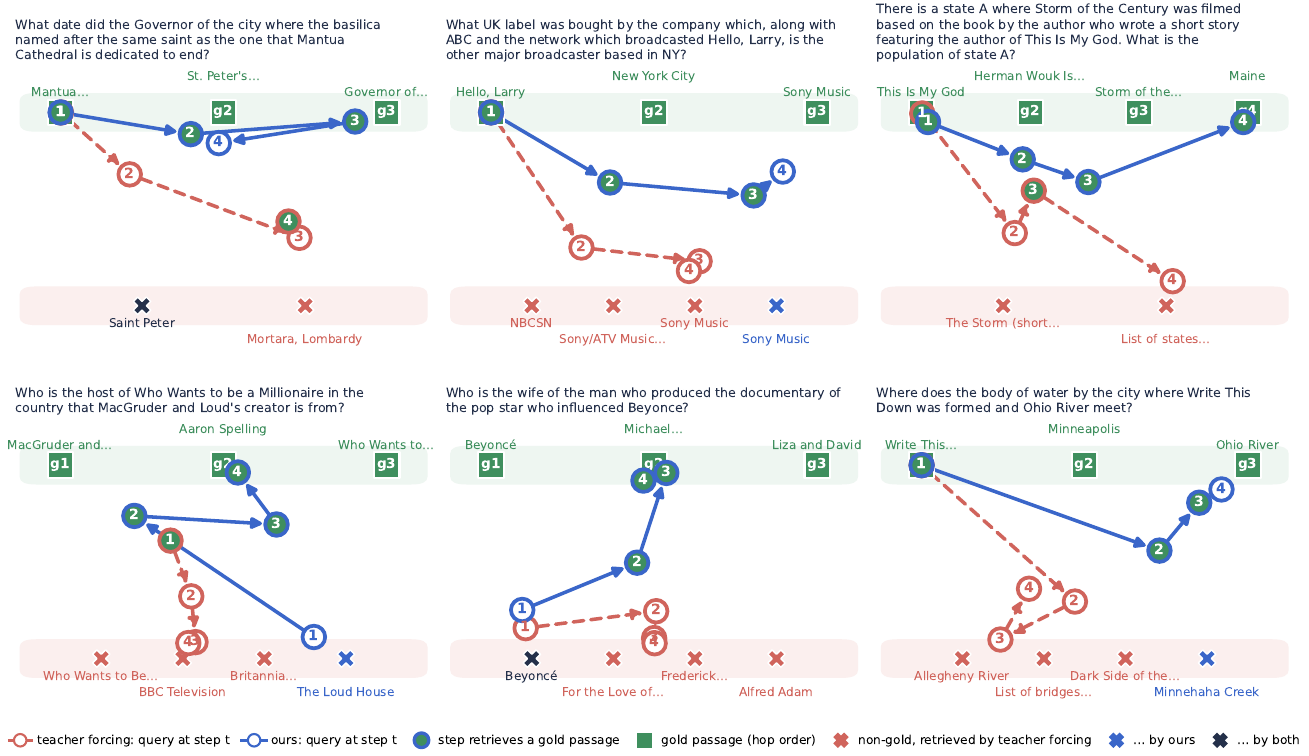}
  \caption{Query trajectories of the two retrievers on six MuSiQue questions,
  each on its own chain. Gold passages sit on the top row in hop order and the
  non-gold passages retrieved by either chain on the bottom row. The query
  vector at step $t$ is drawn at the mean of these positions weighted by the
  softmax of its cosine to each passage at the training temperature
  $\tau=0.05$, so it sits next to the passages it points at; filled nodes are
  steps that retrieve a gold passage.}
  \label{fig:traj}
\end{figure*}

\subsubsection{Where the Query Points}
Figure~\ref{fig:traj} shows the same effect in the embedding space. We place
each query vector by its similarity to the gold passages of the question and
to the non-gold passages the two chains retrieve. The six questions share no
gold passage and are the clearest of the 50 three- and four-hop questions on
which our retriever finds every gold passage and the teacher-forced one
misses at least one, ranked by the gap between the two retrievers in the
softmax weight on the passage each should retrieve next and on non-gold
passages. Once the teacher-forced chain
retrieves a non-gold passage, its next query points at the non-gold passages
and stays there, so the following steps retrieve more of them. Our query moves
along the gold row hop by hop, and when it starts near non-gold passages
(the fourth and fifth questions), it moves up to the gold row within one or
two steps.

\subsection{RQ4: What in the Visited States Matters}
\label{sec:rq4}

\subsubsection{Synthetic, Fixed and Refreshed States}

A cheaper explanation is that any contaminated history would do, as in
training language models with distractor passages
\cite{yoran2024robust,zhang2024raft}. Table~\ref{tab:decomp} separates the
candidate explanations. Every row keeps half of the training states as gold
states and changes the other half, under the same budget and recipe.
Synthetic contamination inserts passages from the frozen encoder's top 100
at random positions into gold histories, with the joint distribution of the
number of wrong and gold passages matched to the visited states in aggregate
(not per question or per epoch), and needs no roll-in; each synthetic state
starts from a teacher-forced state with the drawn number of gold passages.
It is either built once or resampled at every epoch. The
visited half is either sampled once from a single roll-in of the
teacher-forced encoder and kept for all epochs (the w/o refresh condition of
Table~\ref{tab:ablation}), resampled at every epoch from that same roll-in,
or refreshed with a new roll-in at every epoch.

\begin{table}[t]
  \centering
  \caption{What in the visited states matters on MuSiQue; ``new states'' marks a half that changes at every epoch. All
rows are trained with the same seed and recipe.}
  \label{tab:decomp}
  \setlength{\tabcolsep}{3pt}
  \begin{tabular}{lcccc}
    \toprule
    Other half & Roll-in & New states & R@5 & C@5 \\
    \midrule
    Gold (teacher forcing) & -- & -- & \cellcolor{heat!3}69.9 & 44.6 \\
    \midrule
    Synthetic & none & \xmark & \cellcolor{heat!24}72.1 & \cellcolor{heat!14}47.9 \\
    Synthetic & none & \cmark & \cellcolor{heat!14}71.9 & \cellcolor{heat!7}47.1 \\
    Visited & once & \xmark & \cellcolor{heat!7}71.8 & \cellcolor{heat!24}48.4 \\
    Visited & once & \cmark & \cellcolor{heat!7}71.8 & \cellcolor{heat!3}47.0 \\
    Visited (ours) & every epoch & \cmark & \cellcolor{heat!38}\textbf{74.2} & \cellcolor{heat!38}\textbf{50.5} \\
    \bottomrule
  \end{tabular}
\end{table}

Table~\ref{tab:decomp} shows three effects. First, contamination itself helps: synthetic
contamination raises C@5 from 44.6 to 47.9, about half of the gap to on-policy
state training. Second, visited states collected once are worth little more
than synthetic contamination (48.4 against 47.9), so roll-in states have no
special value as such. Third, the remaining gain comes from refreshing with
the current encoder, beyond the fresh samples it provides: resampling the
synthetic contamination at every epoch does not help (47.1), and neither
does resampling the visited states of one roll-in at every epoch (47.0),
while refreshing them with the current encoder raises C@5 to 50.5. On the
fixed states, the resampled condition also recovers no better than the
visited states collected once (32.5\% against 32.2\% after one wrong
passage, against 35.8\% when refreshed). The encoder gains from
training on the states it currently reaches, which move as it learns.

\subsubsection{Shared Negative Pools}
Because the negative pools follow the state text, a visited state also
brings other negatives than a gold state, and this could explain part of the
gain. We therefore give every state of a question the same pool, the frozen
encoder's top 100 for the question alone, and retrain both regimes
(Appendix~\ref{app:sharedneg}). On-policy state training still improves
C@5 over teacher forcing, by 3.8 points against 5.9 with per-state pools, and its
next-step success after a wrong passage remains higher on the fixed states
(32.8\% against 26.1\% after one, 28.7\% against 20.8\% after two or more).
Most of the gain therefore comes from the histories, and a smaller part
from the negatives they induce.

%% file: sections/conclusion.tex
\section{Conclusion}
\label{sec:conclusion}

Chain embedding retrievers are trained on gold histories and run on their
own retrievals. We showed that this gap is large in practice: by the third
step, about half or more of the states a trained chain retriever visits
contain one of its own mistakes, and these mistakes cost a teacher-forced
encoder far more than the difficulty of the step explains. Training on the
states the encoder visits, labeled with the earliest missing gold passage,
changes this without a new architecture, objective or language model. The
gain appears where the mismatch predicts, in recovery from contaminated
histories, it depends on refreshing the visited states as the encoder
learns, and it carries over to chains longer than those seen in training.

These results have three implications within the setting we study. First,
the distribution of training states is a design choice for chain retrievers
in addition to negatives and targets, and reporting how often test-time
states leave that distribution would make results easier to interpret.
Second, retrievers inside agents and reinforcement learning pipelines face
the same mismatch, and dense supervision on visited states is a cheap
complement to outcome rewards. Third, a trained chain embedding retriever,
which needs no language model at retrieval time, is a strong baseline for
graph-based and agentic retrieval for complex questions.

%% file: sections/appendix.tex
\section{Training Details}
\label{app:details}

\subsection{Encoder}
All retrievers start from Qwen3-Embedding-0.6B \cite{zhang2025qwen3embedding}.
Queries are the serialized states $x(s)$ with the model's retrieval
instruction, and passages are encoded without an instruction. Both use
last-token pooling and $\ell_2$ normalization. Document embeddings are
computed once with the frozen model in half precision. The query encoder is
adapted with LoRA \cite{hu2022lora} of rank 16 ($\alpha=32$, dropout 0.05) on
the query, key, value and output projections of every attention layer.

\subsection{States and Serialization}
Of the 2,000 MuSiQue training questions, 200 are used for checkpoint
selection. For a training question with $n$ gold passages, teacher forcing
builds $n$ states; each draws a size $k$ uniformly from $\{0,\dots,n-1\}$,
takes a random subset of $k$ gold passages as the history in chain order, and
targets the earliest gold passage not in it. Chain order follows the question
decomposition of MuSiQue. This gives about 4,700 training
states (4,700 for teacher forcing and 4,691 for on-policy state training,
whose two halves are rounded), the budget of every regime. On-policy state training keeps a fixed
random half of them (about 2,350 states) and fills the other half
by sampling uniformly from the deduplicated visited states, so questions
whose chains stay incomplete for more steps contribute more visited states.
The state text $x(s)$ lists the retrieval instruction, the question and the
passages of the history in retrieval order, each as its index, title and
text, or marks an empty history. Passages are not shortened, and states stay
far below the encoder's limit of 32,768 tokens, so no input is truncated.

Algorithm~\ref{alg:ops} summarizes the procedure.

\begin{algorithm}[h]
  \caption{On-policy state training}
  \label{alg:ops}
  \begin{algorithmic}[1]
    \Require training questions $\mathcal{Q}$ with gold sets $G(q)$; frozen index $\{h(d)\}$; teacher-forced encoder $\theta_{\mathrm{tf}}$ (roll-in only); chain length $T$; state budget $M$; epochs $K$
    \State $\mathcal{D}^{\mathrm{gold}} \gets M$ teacher-forced states $(q, E)$ with $E$ a random gold subset in chain order
    \State initialize LoRA parameters $\theta_1$ from scratch; $\phi \gets \theta_{\mathrm{tf}}$ \Comment{roll-in policy for epoch 1}
    \For{$k = 1, \dots, K$}
      \State $\mathcal{V} \gets \emptyset$
      \For{each $q \in \mathcal{Q}$} \Comment{roll in}
        \State $E \gets ()$
        \For{$t = 1, \dots, T$}
          \State $s \gets (q, E)$; \textbf{if} $R(s) \neq \emptyset$ \textbf{then} $\mathcal{V} \gets \mathcal{V} \cup \{s\}$
          \State $e_t \gets \arg\max_{d \notin E} \sigma_\phi(s, d)$; append $e_t$ to $E$
        \EndFor
      \EndFor
      \State $\mathcal{D}^{\pi}_k \gets M$ states sampled from $\mathcal{V}$ after removing duplicates
      \For{each state $s \in \mathcal{D}^{\mathrm{gold}} \cup \mathcal{D}^{\pi}_k$} \Comment{label}
        \State $a^*(s) \gets$ earliest gold passage in $R(s)$; mask $R(s) \setminus \{a^*(s)\}$
        \State $N(s) \gets$ frozen top-100 of $x(s)$ minus $G(q)$
      \EndFor
      \State $\theta_{k+1} \gets$ minimize $\sum_{s} \ell_\theta(s)$ (Eq.~\ref{eq:loss}) over $\mathcal{D}^{\mathrm{gold}} \cup \mathcal{D}^{\pi}_k$, starting from $\theta_k$ \Comment{update}
      \State $\phi \gets \theta_{k+1}$ \Comment{refresh: next roll-in uses the updated encoder}
    \EndFor
    \State \Return the checkpoint with the best validation C@5
  \end{algorithmic}
\end{algorithm}

\subsection{Optimization}
Each batch contains 16 states. Every state contributes its target and 15
negatives sampled from its frozen top-100 pool, and all 256 candidates of the
batch serve as negatives for every state, with the state's own missing gold
passages masked. We use $\tau=0.05$, AdamW with learning rate
$3\times10^{-4}$ and no weight decay, and a linear warmup over the first 5\% of
steps followed by cosine decay. Retrieval runs for $T=4$ steps during training and on
MuSiQue, and for $T=8$ steps on the long-chain benchmarks. Training lasts at
most 10 epochs, and the checkpoint is selected by chain C@5 on the 200 validation
questions with a patience of 3 epochs. In the first epoch the
visited states come from a roll-in of the teacher-forced encoder; from the
second epoch on they are collected by the current encoder before any update
of that epoch. Ablations and teacher-forced retrievers use the same recipe
and the same number of training states. Table~\ref{tab:main} averages seeds 42 to 46; every other experiment trains
each retriever once with seed 13.

\section{Shared Negative Pools}
\label{app:sharedneg}

In this control, every training state of a question, teacher-forced or
visited, draws its 15 negatives from one pool: the frozen encoder's top 100
for the question alone, with the gold passages removed. Sampling, in-batch
negatives (which still come from the other states of the batch) and all
other settings are unchanged. Next-step
success is measured on the fixed states of Section~\ref{sec:rq3}, by the
number of wrong passages in the history. With per-state
pools, the pool of a state with a wrong passage in its history shares 43\% of
its passages with the question pool, against 57\% for a clean history.

\begin{table}[h]
  \centering
  \caption{Per-state and shared negative pools on MuSiQue, all trained with the same seed.}
  \label{tab:sharedneg}
  \setlength{\tabcolsep}{3pt}
  \begin{tabular}{llc|ccc}
    \toprule
    & & & \multicolumn{3}{c}{Next-step success} \\
    Pools & Training & C@5 & 0 & 1 & 2+ \\
    \midrule
    \multirow{2}{*}{Per state} & Teacher forcing & \cellcolor{heat!7}44.6 & \cellcolor{heat!14}67.0 & \cellcolor{heat!7}25.0 & \cellcolor{heat!7}14.5 \\
    & On-policy & \cellcolor{heat!38}\textbf{50.5} & \cellcolor{heat!38}\textbf{69.0} & \cellcolor{heat!38}\textbf{35.8} & \cellcolor{heat!38}\textbf{29.2} \\
    \midrule
    \multirow{2}{*}{Shared} & Teacher forcing & \cellcolor{heat!14}45.0 & \cellcolor{heat!7}65.4 & \cellcolor{heat!14}26.1 & \cellcolor{heat!14}20.8 \\
    & On-policy & \cellcolor{heat!24}\textbf{48.8} & \cellcolor{heat!24}\textbf{67.9} & \cellcolor{heat!24}\textbf{32.8} & \cellcolor{heat!24}\textbf{28.7} \\
    \bottomrule
  \end{tabular}
\end{table}

\section{Mixing Ratio}
\label{app:mix}

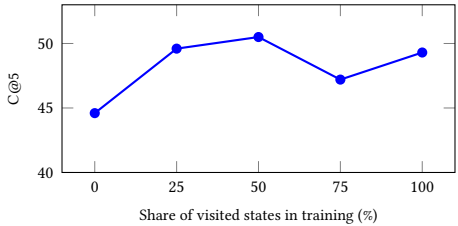
\begin{figure}[h]
  \centering
  \begin{tikzpicture}
    \begin{axis}[
      width=0.8\columnwidth, height=3.8cm,
      xlabel={Share of visited states in training (\%)}, ylabel={C@5},
      xtick={0,25,50,75,100}, ymin=40, ymax=53,
      tick label style={font=\scriptsize}, label style={font=\scriptsize},
      ]
      \addplot[thick, blue, mark=*, mark size=1.5pt] coordinates {(0,44.6) (25,49.6) (50,50.5) (75,47.2) (100,49.3)};

    \end{axis}
  \end{tikzpicture}
  \caption{Mixing ratio between gold and visited states on MuSiQue, all trained
  with the same seed; 0\% is teacher forcing on gold states.}
  \label{fig:mix}
\end{figure}

The share of visited states is $1-\lambda$ in Eq.~\eqref{eq:objective}, and
Figure~\ref{fig:mix} varies it. Every share of visited
states improves on teacher forcing, and visited states alone recover about
80\% of the gain of the best mixture, which again places the effect in the
states the encoder reaches. The curve is not monotone, and the gold half
adds a smaller and less certain gain: the equal mixture is best, by 1.2 points
over visited states alone. We keep it because it preserves supervision on
histories that the current encoder rarely reaches.

\section{HippoRAG 2 on Reduced Corpora}
\label{app:pool}

HippoRAG~2 and G-Retriever are run on a reduced corpus for each long-chain
benchmark, a simpler setting than the full corpora. It contains all gold
passages and the top 100 passages of every question under the frozen
encoder (27,744 passages for HotpotQA-Long and 19,257 for 2Wiki-Long).
HippoRAG~2 uses deepseek-flash for extraction and filtering, and the other
retrievers are re-evaluated on the same corpora.
Table~\ref{tab:pool} reports the results. The triple filter keeps no fact
for 87 HotpotQA-Long and 48 2Wiki-Long questions, for which HippoRAG~2 falls
back to dense retrieval; on chains of three to five hops, a single round of
fact retrieval rarely reaches the later passages.

\begin{table}[h]
  \centering
  \caption{Results on the reduced corpora (R@10 / C@10); ours is the five-seed mean of Table~\ref{tab:main}.}
  \label{tab:pool}
  \begin{tabular}{lcc}
    \toprule
    Method & HotpotQA-Long & 2Wiki-Long \\
    \midrule
    Dense & 31.9 / 1.0 & 34.5 / 2.0 \\
    HippoRAG 2 & 33.4 / 1.3 & 35.7 / 2.3 \\
    On-policy state training & \textbf{64.2} / \textbf{38.6} & \textbf{69.2} / \textbf{36.8} \\
    \bottomrule
  \end{tabular}
\end{table}

\section{Long-Chain Benchmarks}
\label{app:data}

\subsection{Chains}
2Wiki-Long uses the evidence triples of 2WikiMultiHopQA as a graph over
Wikidata entities, keeping only functional relations so that every hop has a
single answer. Chains are paths of three to five hops in which every hop is
supported by the paragraph of its subject entity, no entity repeats, and
degenerate relation pairs (for example, the spouse of a mother) are excluded.
Comparison questions map three or four entities through one relation to a
person and then to a date, and ask which entity's related person has the
earliest or latest date. HotpotQA-Long starts from the two pages of an
original bridge question and follows hyperlinks in the 2017 Wikipedia
abstracts to reach three, four or five pages. Hops go only to named-entity
pages with in-degree at most 300, through sentences with fewer than five
links.

\subsection{Questions and Filters}
A language model (deepseek-flash) writes one question per chain. The question may name only
the starting entity, or the compared entities, and uses every relation in
order. A question is kept only if it names no intermediate entity or answer,
is answered correctly from the shuffled gold passages, is judged to have a
unique answer, and is rated 4 or 5 for naturalness, all judged by the same
model. A model of a different family (gpt-5.4-mini) then re-checks
answerability and uniqueness; 62\% of the 2Wiki-Long
pool and 98\% of the HotpotQA-Long pool pass. Questions whose final gold
passage (for comparisons, all gold passages) appears in the top 10 of
question-only BM25 over the final corpus are removed. Genealogy questions
are capped at 40\% of 2Wiki-Long, and question types are sampled evenly.
HotpotQA-Long has 100 test questions each of three, four and five hops;
2Wiki-Long has 170, 45 and 12 chains of three, four and five hops and 73
comparison questions. The prompts are released with the benchmarks. Without
passages, the reader of Section~\ref{sec:main} answers 14.3\% of the
HotpotQA-Long and 14.7\% of the 2Wiki-Long test questions, so the benchmarks
require retrieval. Table~\ref{tab:examples} shows one test question of four
types.

\begin{table*}[t]
  \centering\small
  \caption{Example test questions of the long-chain benchmarks with their gold chains.}
  \label{tab:examples}
  \begin{tabular}{p{0.11\textwidth}p{0.45\textwidth}p{0.38\textwidth}}
    \toprule
    Type & Question (answer) & Gold passages in chain order \\
    \midrule
    HotpotQA-Long, 3 hops & Who voices the title character of the animated television series that features Kuvira as a fictional character? (Janet Varney) & Kuvira $\to$ The Legend of Korra $\to$ Korra \\
    HotpotQA-Long, 5 hops & Hagen Site sits in a county whose seat was headquarters for a rail division that reached a town on a river; how long is that river? (692 mi) & Hagen Site $\to$ Dawson County, Montana $\to$ Glendive, Montana $\to$ Livingston, Montana $\to$ Yellowstone River \\
    2Wiki-Long, 4 hops & Where was the father of the maternal grandmother of Prince Andrew of Greece and Denmark born? (Hildburghausen) & Prince Andrew of Greece and Denmark $\to$ Olga Constantinovna of Russia $\to$ Princess Alexandra of Saxe-Altenburg $\to$ Joseph, Duke of Saxe-Altenburg \\
    2Wiki-Long, comparison & Among The Lambeth Walk, Dikshul, and Ezhu Rathrikal, which entity has the composer with the latest date of death? (Ezhu Rathrikal) & The Lambeth Walk, Noel Gay, Dikshul, Pankaj Mullick, Ezhu Rathrikal, Salil Chowdhury \\
    \bottomrule
  \end{tabular}
\end{table*}

\subsection{Corpora}
Each corpus contains every gold passage of the candidate pool, then the BM25
nearest neighbors of each gold passage, taken round-robin over gold passages,
then random passages from the source collection. HotpotQA-Long has 554,812
passages, of which 328,648 are nearest neighbors; 2Wiki-Long has 220,000
passages, of which 95,880 are nearest neighbors. A development split of 100
questions per benchmark is built with the same construction. Test
questions come from both original splits of the source datasets, which is
possible because the benchmarks are used only for evaluation.

\section{Additional Results}
\label{app:more}

\subsection{Question Answering}
Table~\ref{tab:qa2} extends Table~\ref{tab:qa} with teacher forcing on the same
encoder and with a second reader, qwen-flash, under the same prompt. The
passages of on-policy state training give 5.5 more EM and 5.7 more F1 than
those of teacher forcing with deepseek-flash, and 2.5 and 3.5 more with
qwen-flash. The ranking of the retrievers is the same under both readers.

\begin{table}[!ht]
  \centering
  \caption{Question answering on MuSiQue with two readers; trained retrievers use the same seed.}
  \label{tab:qa2}
  \begin{tabular}{lcc|cc}
    \toprule
    & \multicolumn{2}{c|}{deepseek-flash} & \multicolumn{2}{c}{qwen-flash} \\
    Passages from & EM & F1 & EM & F1 \\
    \midrule
    Dense & 25.9 & 36.3 & 18.8 & 28.8 \\
    G-Retriever & 21.2 & 29.2 & 15.7 & 24.3 \\
    HippoRAG 2 & 28.2 & 39.4 & 21.3 & 30.8 \\
    LinearRAG & 22.3 & 30.5 & 17.3 & 26.0 \\
    Q-RAG & 32.4 & 42.0 & 25.5 & 35.4 \\
    Search-R1 & 35.5 & 44.8 & 27.7 & 37.2 \\
    Teacher forcing & 38.6 & 48.0 & 29.8 & 40.0 \\
    On-policy state training & \textbf{44.1} & \textbf{53.7} & \textbf{32.3} & \textbf{43.5} \\
    \bottomrule
  \end{tabular}
\end{table}

\subsection{Cost}
\label{app:cost}
Training adds a roll-in per epoch to teacher forcing: collecting
$\mathcal{D}^{\pi}_k$ takes $T$ query encodings and nearest-neighbor searches
per training question, plus one negative lookup per sampled state, and no
passage is re-encoded. The first-epoch roll-in uses a teacher-forced
encoder, so the full pipeline costs one teacher-forced training run more
than teacher forcing alone. Teacher forcing trains for 10
epochs in 37 minutes; on-policy state training stops early after 7 epochs
and takes 60 minutes, 32 of them for the roll-ins. Both regimes schedule the
same 2,940 optimizer steps. At test time the trained encoder runs the same
greedy chain over the same index.

Table~\ref{tab:cost} compares the cost of retrieval on MuSiQue. Our retriever
adds 4.6M trained parameters to the 0.6B encoder, calls no language model and
needs only the passage embeddings as a dense index; one question takes 0.10\,s for
four chain steps. The graph-based systems call a language model over every
passage to build their index, and Search-R1 runs a 7B language model for
about four generation turns per question.

\begin{table}[!ht]
  \centering
  \caption{Cost of retrieval on MuSiQue (11,656 passages).}
  \label{tab:cost}
  \setlength{\tabcolsep}{3pt}
  \begin{tabular}{lccc}
    \toprule
    Method & Parameters & LLM calls / question & Index \\
    \midrule
    BM25 & -- & 0 & inverted \\
    Dense & 0.6B & 0 & dense \\
    G-Retriever & 0.6B & 0 & LLM graph \\
    HippoRAG 2 & 0.6B + LLM & 1 & LLM graph \\
    LinearRAG & 0.6B + NER & 0 & entity graph \\
    MDR & 0.25B & 0 & dense \\
    Q-RAG & 1.1B & 0 & dense \\
    Search-R1 & 7.6B + 0.6B & 4.0 & dense \\
    Ours & 0.6B & 0 & dense \\
    \bottomrule
  \end{tabular}
\end{table}

Building the HippoRAG~2 and G-Retriever graphs takes 23.3k language model
calls with 14.6M tokens on MuSiQue. Dense retrieves a question
in 0.017\,s and our retriever in 0.104\,s.

\subsection{Beam Decoding}
Beam decoding is a cheaper way to recover from an early mistake, so we decode
both matched encoders with beams of two and four. Each beam is a partial chain
scored by the sum of its similarities; at every step each beam expands its top
candidates, the best partial chains are kept, and the passages of the best
final chain are ranked first, followed by the maximum score of every passage
over all encoded states. A beam of one is greedy decoding. Table~\ref{tab:beam}
shows that beams improve teacher forcing by two to three points of complete
coverage but leave a gap of four to six points, and that the greedy on-policy
retriever outperforms the teacher-forced one with a beam of four, which
encodes three times as many queries. Both encoders are those of
Table~\ref{tab:ablation}; the re-encoded index moves greedy C@10 on
HotpotQA-Long by one question.

\begin{table}[h]
  \centering
  \caption{Greedy and beam decoding with the matched encoders of the same seed (queries encoded per question; TF is teacher forcing).}
  \label{tab:beam}
  \setlength{\tabcolsep}{3pt}
  \begin{tabular}{lccc|ccc}
    \toprule
    & \multicolumn{3}{c|}{MuSiQue (C@5)} & \multicolumn{3}{c}{HotpotQA-Long (C@10)} \\
    Decoding & Queries & TF & Ours & Queries & TF & Ours \\
    \midrule
    Greedy & 4 & 44.6 & \textbf{50.5} & 8 & 22.7 & \textbf{28.3} \\
    Beam 2 & 7 & 46.4 & \textbf{50.6} & 15 & 25.7 & \textbf{29.7} \\
    Beam 4 & 13 & 46.3 & \textbf{50.7} & 29 & 24.3 & \textbf{30.3} \\
    \bottomrule
  \end{tabular}
\end{table}

\subsection{Recovery on the Long-Chain Benchmarks}
Table~\ref{tab:longprobe} repeats the fixed-state probe of
Figure~\ref{fig:diagnosis}(b) on the states of the teacher-forced chains of
the long-chain benchmarks ($T=8$). As on MuSiQue, the three encoders are close
when the history is clean, and the on-policy encoder succeeds about twice as
often once the history holds two or more wrong passages.

\begin{table}[!ht]
  \centering
  \caption{Next-step success on the teacher-forced states of the long-chain benchmarks; all encoders use the same seed.}
  \label{tab:longprobe}
  \setlength{\tabcolsep}{3.5pt}
  \begin{tabular}{lccc|ccc}
    \toprule
    & \multicolumn{3}{c|}{HotpotQA-Long} & \multicolumn{3}{c}{2Wiki-Long} \\
    Wrong passages & 0 & 1 & 2+ & 0 & 1 & 2+ \\
    \midrule
    Teacher forcing & \textbf{56.0} & 25.1 & 7.4 & 68.2 & 22.3 & 8.1 \\
    w/o refresh & 55.8 & 31.1 & 14.3 & 70.3 & 26.9 & 20.5 \\
    On-policy state training & 55.2 & \textbf{31.7} & \textbf{16.7} & \textbf{74.1} & \textbf{36.9} & \textbf{21.0} \\
    \bottomrule
  \end{tabular}
\end{table}

\subsection{Error Propagation}
\begin{figure}[!ht]
  \centering
  \includegraphics[width=\columnwidth]{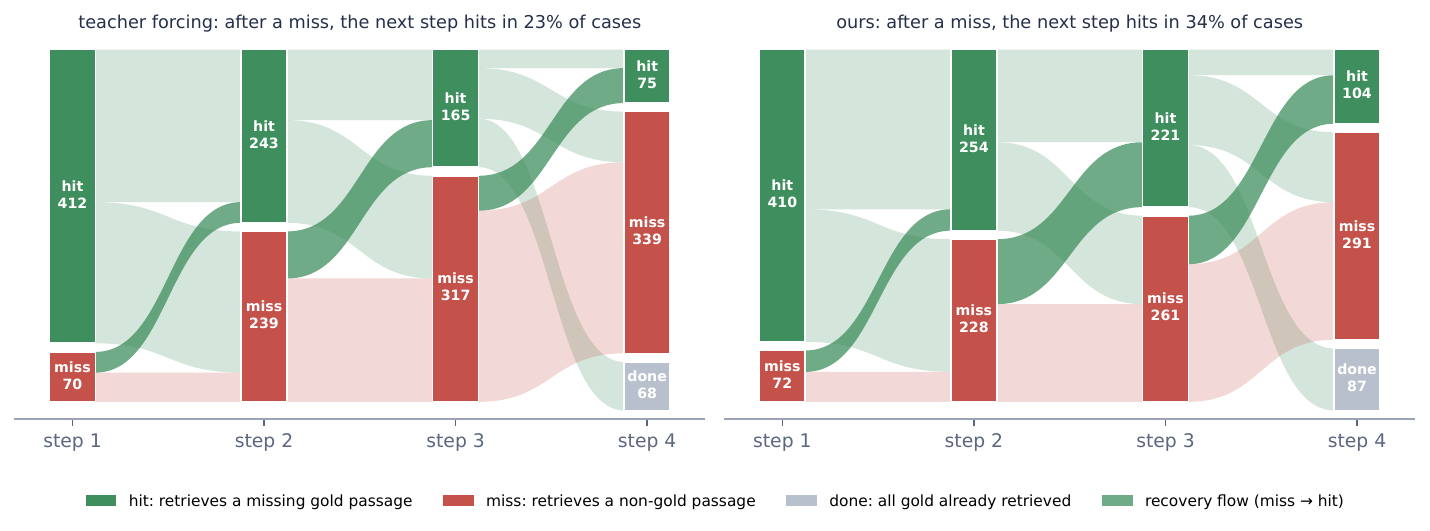}
  \caption{Step-by-step outcomes on three- and four-hop MuSiQue questions
  (shared seed); dark bands are recoveries from a miss to a hit.}
  \label{fig:flow}
\end{figure}

Figure~\ref{fig:flow} follows the 482 three- and four-hop MuSiQue questions
along each retriever's own chain. After a step that retrieves a non-gold
passage, the next step retrieves a missing gold passage in 23.0\% of the
cases under teacher forcing and in 33.7\% under on-policy state training,
and the chain is complete after four steps for 68 and 87 questions.

\subsection{Where Contaminated Queries Point}
\begin{figure}[!ht]
  \centering
  \includegraphics[width=0.95\columnwidth]{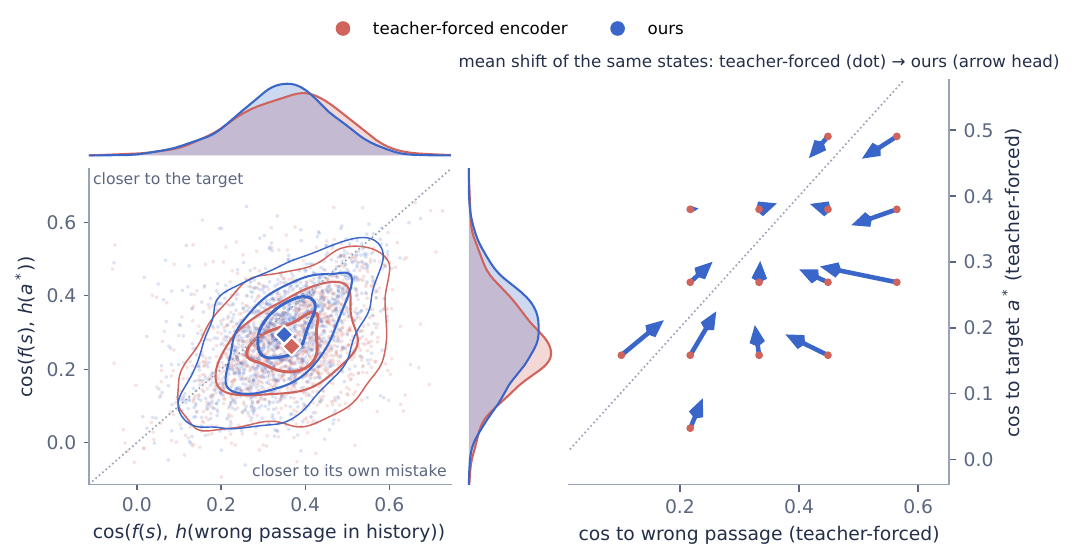}
  \caption{Cosine of contaminated queries to the wrong passage in their history and to the target.}
  \label{fig:attract}
\end{figure}

On the 1,358 contaminated states (Figure~\ref{fig:attract}; shared seed), the query is
closer to the target than to the wrong passage in its history in 20.3\% of
the states for teacher forcing and in 30.6\% for on-policy state training.

